# CH-Go: Online Go System Based on Chunk Data Storage

Hui LU
*Computer School*
*Sichuan University*
Chengdu, China
2018141501149@stu.scu.edu.cn

Chuan LI *
*Computer School*
*Sichuan University*
Chengdu, China
lcharles@scu.edu.cn

Yiming YANG
*School of Computing*
*National University of Singapore*
Singapore, Singapore
e0920761@u.nus.edu

Cheng LI
*Computer School*
*Sichuan University*
Chengdu, China
996515149@qq.com

Ashraful Islam
*Computer School*
*Sichuan University*
Chengdu, China
ash_512@outlook.com

*Abstract*—The training and running of an online Go system require the support of effective data management systems to deal with vast data, such as the initial Go game records, the feature data set obtained by representation learning, the experience data set of self-play, the randomly sampled Monte Carlo tree, and so on. Previous work has rarely mentioned this problem, but the ability and efficiency of data management systems determine the accuracy and speed of the Go system. To tackle this issue, we propose an online Go game system based on the chunk data storage method (CH-Go), which processes the format of 160k Go game data released by Kiseido Go Server (KGS) and designs a Go encoder with 11 planes, a parallel processor and generator for better memory performance. Specifically, we store the data in chunks, take the chunk size of 1024 as a batch, and save the features and labels of each chunk as binary files. Then a small set of data is randomly sampled each time for the neural network training, which is accessed by batch through yield method. The training part of the prototype includes three modules: supervised learning module, reinforcement learning module, and an online module. Firstly, we apply Zobrist-guided hash coding to speed up the Go board construction. Then we train a supervised learning policy network to initialize the self-play for generation of experience data with 160k Go game data released by KGS. Finally, we conduct reinforcement learning based on REINFORCE algorithm. Experiments show that the training accuracy of CH-Go in the sampled 150 games is 99.14%, and the accuracy in the test set is as high as 98.82%. Under the condition of limited local computing power and time, we have achieved a better level of intelligence. Given the current situation that classical systems such as GOLAXY are not free and open, CH-Go has realized and maintained complete Internet openness.

## I. Introduction

Human-machine play is closely related to artificial intelligence [1], and it is one of the most important ways to testify computational intelligence. This is especially true for Go, which has always been regarded as the most challenging classic game in artificial intelligence due to its vast search space and difficulty evaluating chess positions and moves [2]. Every advance in the Go intelligent system since the 1990s is bound to be accompanied by a renowned man-machine battle.

For a long time, Go has been regarded as the embodiment of the highest difficulty among all human chess games, so many famous teams worldwide have made active attempts to solve problems in the domain of Go. Among them, the open source system, Pachi, based on the traditional heuristic Monte Carlo tree search is released in 2012. It is characterized by a modular structure and a small and refined code base [3], but deep learning is not introduced in their work. ELF OpenGo [4], released by Facebook AI Research team (FAIR), implements its early achievements Darkforest [5] as well as the excellent AlphaGo Zero [6] and AlphaZero [7] algorithms proposed by the DeepMind team, et al. There are also some non-open systems such as AlphaGo [8], the first Go intelligence developed by the Google DeepMind team which defeats the human Go world champion, integrated deep learning, reinforcement learning and innovatively applied Monte Carlo tree search for the first time. However, these open source systems have the disadvantages of requiring downloading and configuration, complicated operations, and inability to access online. Therefore, CH-Go (Chunk data storage-based Go system) aims to build a simple and easy-to-use online Go human-machine game system.

However, most of the existing systems require massive computing power and huge investment, and there is insufficient consideration for how to implement artificial intelligence systems with limited resources. In response to that, we propose a system construction method based on the chunk data management strategy to solve the problems of large amount of Go record data, slow reading speed, and large memory space. Based on the pre-processed game records, our method analyzes 160k Go data, that is, millions of moves in a unique binary format of Numpy with a block size of 1024 according to features and labels, and uses the effective generator yield function when accessing. With the progress of supervised learning neural network training, data are generated and returned in batches.

In addition, we apply Zobrist [9]-guided hash coding when constructing the Go board, and design 11 feature planes as the input of the neural network. When training the neural network, considering the different characteristics of Go moves in different stages, the adaptive gradient descent Adadelta [10] method is used to optimize the neural network. Then, in the

---

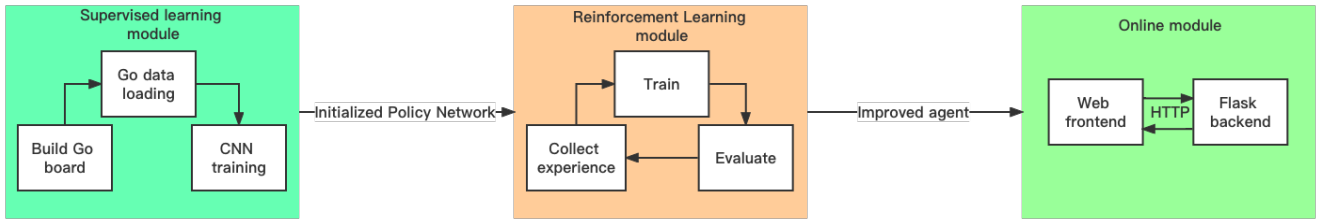

Fig. 1. System structure diagram

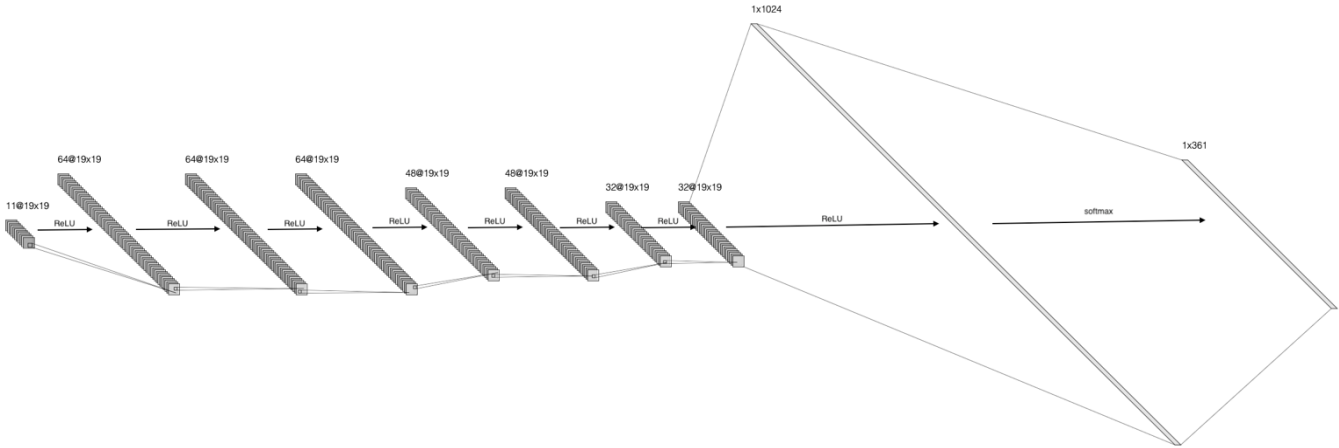

Fig. 2. Deep Convolutional Neural Network Mode

part of reinforcement learning, we adopt the clipping probability distribution and random sampling to select actions to improve the stability of the model and use the REINFORCE algorithm based on the policy gradient in the training. At last, we build a lightweight Flask application with web front-end to run the instance agent, which is deployed on the server to provide a platform for Go enthusiasts to play human-machine games. To summarize, our contributions are three-fold:

1) We propose a truncated storage module for the Go system. A large amount of Go score data is stored in blocks, so the effective yield method can be used to provide data in batches according to the training process to improve storage and access efficiency.

2) We introduce a Zobrist-guided hash coding module to facilitate the building of Go board, which significantly speeds up the game and improves the training effect with the limited computing power.

3) We design 11 feature planes as the input of the neural network, enjoying the merits of high training speed and significant performance. Our system achieves 98% accuracy in sampled dataset.

## II. THE ARCHITECTURE OF CH-GO

The system first builds the basic Go game and downloads the game records from the KGS Go server. Then the chess format decoder and input encoder are generated to transform the data into the required input format. After that, the system conducts supervised training and reinforcement learning with the REINFORCE algorithm. Finally, the Web application is deployed on the server. As shown in Fig. 1, our proposed system consists of three modules, namely supervised learning module, reinforcement learning module, and online module.

### A. Supervised learning module

The supervised learning module mainly includes building Go boards, loading the game records downloaded from KGS (Kiseido Go Server), and a deep convolutional neural network training module.

When building Go boards, in addition to realizing the most important actions in Go, such as moving and raising, it is also necessary to consider many other vital concepts in Go, such as ladder, self-Atari and ko. To detect ko, the game's entire history needs to be checked, which is computationally expensive. In order to reduce the cost of storage and computation, we adopt Zobrist-guided hash code to improve speed.

Then the system needs to conduct a crawler program, downloading online game records from KGS in batches and sampling the specified number of games as required. After that, the system processes the game record data, creates features and labels, and saves the results locally in the form of NumPy arrays in blocks. To avoid making a vast NumPy array, the system also builds a Go data generator to provide a minibatch of features and labels for each neural network training. In addition, the python multiprocessing library is applied to map the loaded data workload to multiple CPUs to facilitate Go data processing in a highly parallel mode. In this paper, we map it to 8 CPUs to avoid memory overflow and other troubles while considerably speeding up data processing. In the training module, a deep convolutional neural network is generated as shown in Fig. 2, which has 9 layers. The network's input is a vector of size 19*19*11, then transformed into a vector of size 25*25 by using zero padding in the first hidden layer. And the input is convolved with K filters with kernel size 7 * 7 and stride set to 1. As for the remaining 2 to 7 hidden layers, the size of the vector becomes 23*23 through zero padding, then convolving with K filters with a kernel size of 5*5 and a stride of 1. After that, it is connected with a dense layer of size 1024. The system applies ReLU for the first 8 layers and the SoftMax function in the last layer, outputting the probability distribution of the 361 classes of the Go board.

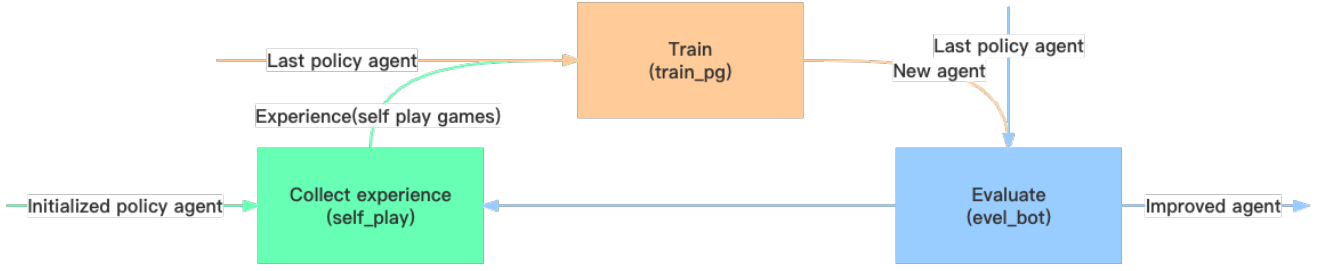

Fig. 3. Reinforcement learning module structure diagram

In addition, the categorical cross-entropy loss function and adaptive gradient descent Adadelta are used to find the minimum value. This method does not set the global learning rate and differs from Adagrad whose learning rate is aggressive and monotonically decreasing. Adadelta restricts the cumulative window of past gradients:

$$G \leftarrow \gamma G + (1-\gamma) \partial W \qquad (1)$$

Where $W$ is the weight vector, $G$ is a diagonal matrix, $G_{i,i}$ notes the sum of squares of gradients $W_i$ received so far and $\gamma$ is the size of window, generally set to 0.9.

Besides, the size of the mini-batch also affects the performance of the model. Generally, the size of the mini-batch is the same with or close to the number of categories, but the moves of Go are not completely random, and the probability of particular moves is significantly higher than that of others. For example, there are almost no moves in the four corners. For the imbalance phenomenon, it is impossible to expect the mini-batch size to cover all categories. Therefore, referring to many experimental results, we set the mini-batch size to 128.

### B. Reinforcement learning module

Theoretically speaking, if the Go intelligent system is trained only through supervised learning, it will never be able to outperform the game records provided to the neural network [9]. Therefore, reinforcement learning is necessarily required to improve the system's level. Fig. 3 shows the module structure.

First, the system initializes the games guided by the supervised learning policy network results, performs a large number of self-plays Go games, and records the states of the board, which aims to collect experience. Then CH-Go adopts the REINFORCE algorithm based on the policy gradient and the idea of Monte Carlo and updates the weights of the initialized neural network through training. Evaluations can be performed after each version of training is completed. In the domain of reinforcement learning, we call each version of the Go robot an agent, and let the new agent play against the earlier version of the agent by self-play and use binomial test to evaluate its progress and determine whether the program needs further improvement. Generally, these three processes need to be repeated for many cycles.

### 1) Task Formulation

Many games with perfect information, such as chess, checkers, reversi, backgammon, and Go, can be defined as alternating Markov games [12]. In Go, the game that the system faces are the state space $S$; the legal move that the system can choose is the action space $A(S)$; the reward function $r^i(s)$ describes the rewards obtained by player i and game s. The result of the game $z_t = \pm r(s_T)$ is the final reward from the view of the current player at time step t when the game is over. At the end of each game, if winning it will harvest a score of 1, otherwise it will get a score of -1. The system's strategy p(a | s) is the probability distribution of legal actions $a \in A(s)$. When both players' actions are selected according to the strategy $p$, the value function is the expected result, namely $v^p(s) = E[z_t | s_t = s, a_{t...T} \sim p]$.

### 2) REINFORCE algorithm

REINFORCE is an acronym for REward Increment = Non-negative Factor * Offset Reinforcement * Characteristic Eligibility [13]. It aims to use stochastic gradient ascent to update the parameters, making the harvest continue to rise and maximizing the expected profit. It is a policy gradient algorithm based on the Monte Carlo update method, every time an episode is completed or a round of Go is over to update the agent. Specifically, an empirical Go game is formed by self-play through a supervised learning policy network, which is input into the initial policy network, and the probability distribution of actions is outputted. The agent randomly selects under certain conditions and fills in 1 or -1 into the target vector after completing a game, representing the result of this episode. The agent gets rewards by interacting with the environment. The difference between the probability distribution and the target vector represents the gradient to be followed, which can then be used to update the network weights using gradient descent. When a batch update is completed, the policy network has new weights, and a new probability distribution will be obtained when the game is inputted again. The formula for stochastic gradient ascent is as follows:

$$\theta_{t+1} \doteq \theta_t + \alpha \sum_a \hat{q}(S_t, a, w) \nabla \pi(a|S_t, \theta) \quad (2)$$

Where different representations of q form different algorithms. We replace q with future total returns $G_t$ in REINFORCE, representing the sum of returns after t steps. So, the algorithm formula is as follows:

$$\theta_{t+1} \doteq \theta_t + \alpha G_t \nabla \ln p(A_t|S_t, \theta_t) \quad (3)$$

Where α is the learning rate. If $G_t > 0$, the parameter update will increase the probability of keeping the current state, that is, if the reward is favorable, the probability of this action will be increased, and vice versa. Besides, the greater the reward, the greater the magnitude of the gradient update and the greater the probability increase. The reason for choosing log-probability ln*p* over probability (which is what we should maximize) is that, in general, optimizing log-probability works better than normal probability and the gradient of log-probability is usually easier to scale. Within the probability range 0-1, the variable space of probability *p* is limited and small, but when using logarithmic probability, we have the target (-∞,0 ) with a more extensive "dynamic range" than the original probability space, which makes the log probability easier to compute.

Reinforcement learning processes can be particularly unstable in the early stage of training, where the agent may assign a pretty high probability to chance wins, even though the actions are actually not so effective. Therefore, the system adjusts the probability distribution and adds a particular random component, which enables the exploration of various possibilities of Go games [14].

First, all values are cubed, drastically increasing the distance between more and less likely actions. While the best possible actions should be chosen as much as possible, at the same time, the system should prevent the probabilities of actions from getting too close to 0 and 1. Therefore, the system defines a small positive value ε = 0.000001 [9], and sets all values less than ε to ε and all values greater than 1-ε to 1-ε. Finally, all results are normalized, and we get another probability distribution. In addition, the system randomly samples from this probability distribution, rather than directly selecting the most likely action, because usually the system cannot find the action that stands out from so many options, and sampling also avoids the system repeatedly selecting the same activity, improving the diversity of actions and the stability of the network.

During the experiment, it was found that generating a large number of Go games in one batch is not conducive to the weight update of the REINFORCE algorithm. When the learning rate is very low, the performance of the model is even worse. Experiments show that its iterative progress is faster on a small amount of data. Referring to AlphaGo's use of 10k mini-batches of 128 games, we finally choose to generate 128 self-play Go games each time, and use the REINFRORCE method to train and update the weight of the neural network. Each time a version of the training is completed, it is played against the previous version or the last version before it, and the agent's progress is evaluated. When its winning rate is similar or dominant in 100 games, in order to perform a binomial test, we choose to continue playing 1000 games. After 20 iterations, the obtained agent wins 565 games against the initial strategy network in 1000 games. The binomial test result is 4.14e-05, which is far less than the general standard of 5%, so it can be considered that he has made significant progress.

### C. Online module

The architecture of the online system module is shown in Fig. 4. When the user performs an action in the browser, the web front-end sends the game state through an HTTP POST request to the Flask application, which is responsible for decoding the POST request and passing it to the Go agent instance. In the browser, the client based on the jgoboard library communicates with the server via HTTP.

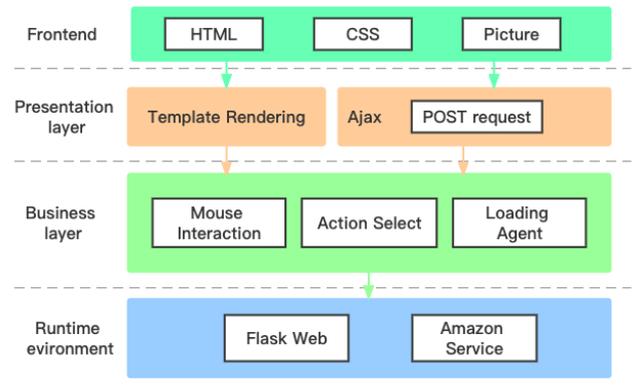

Fig. 4.  Online Module Architecture Diagram

### III. METHODOLOGY

### A. Truncated storage module

Numpy binary files have a lot of advantages such as low memory requirement, high reading speed and suitable for model reloading and migration. Therefore, after pre-processing the downloaded data, the features and labels are saved in the form of Numpy arrays. The storage of massive small files is an important research direction in today's massive data storage technology [15]. We choose the block storage method considering the need to access features repeatedly in small batches at the late-stage neural network. The block size of 1024 is regarded as a batch (1024 is an integral multiple of the mini-batch of neural network training). After block storage, chunk data can not only be combined when using, but can also dynamically load batch data, loading it into memory separately with high flexibility.

When accessing data, we solve the problem of large memory consumption when using sequences to store large amounts of data by applying the mechanism of computing while looping in python generators. Yield is one of the methods to implement generators, and simultaneously assumes the responsibility of accessing data. The difference between yield and return function is that the return function returns a list, while yield only returns one value, so yield dramatically reduces the overhead of accessing data. The pseudo-code of access algorithm to the block data storage of our system is as follows:

| Algorithm 1. Data Block Access |
| --- |
| Input: 160k go data with zip format, |
| Output: the outcome of data storage by chunk |
| Steps: |
| 1. **for** zip_file in file: |
| 2. unzip_data |
| 3. feature, label = process_zip (batch_size = 1024) |
| 4. for feature, label: |
| 5. x = load(feature) |
| 6. y = load(label) |
| 7. while x.shape >= batch_size: |
| 8. x_batch, x = x[:batch_size], x[batch_size:] |
| 9. y_batch, y = y[:batch_size], y[batch_size:] |
| 10. yield x_batch, y_batch |

A generator also calls this method with the batch size of 128. We only access a small batch of data each time calling the function by our algorithm, which effectively avoids the problem of memory leakage due to vast arrays. At the same time, it efficiently provides features and labels for training of the neural network.

*B. Zobrist-guided hash coding module*

Now, most AI for games applies Zobrist hashes to implement transpose tables [16][17]. For example, in the game of Go, different moves are likely to result in the same situation. With Zobrist, a special type of hash table, we can avoid analyzing the same position repeatedly indexed by the position of the board. Precisely, Black and White stones are placed on a 19*19 board. At any time, there are three black, white and empty states at each intersection on the board, generating 3*19*19=1083 hash values. Using these values to represent a single action, we do the XOR operation between the initial hash value of the board and the action to obtain a new board hash value when a move occurs. When capture occurs, we just need to apply that to remove the stone reversely.

When detecting whether ko occurs, we first limit to 10 moves, that is, 5 situations, assuming they are in different order with the same final position. If an explicit loop is used, this same situation will be calculated 5! (=120) times. Then if we need to detect ko within 20 moves, the calculation time required by Zobrist algorithm is only 1/3628800 (=1/10!) of the original ideally.

Therefore, the playing speed will be very fast at any time by the Zobrist-guided algorithm, especially in self-play games, and the efficiency of the system is significantly improved.

*C. Input feature design*

A total of 48 input features in AlphaGo's supervised learning policy network are directly derived from raw representations of game rules [8]. More features mean more cost of computation and time. Therefore, we design a small input feature with 11 feature planes by trade-off effectiveness and efficiency, as shown in Table 1:

TABLE I. INPUT FEATURES FOR NEURAL NETWORKS

| Feature | # of planes | Description |
| --- | --- | --- |
| Liberties | 4 | Number of privileges (empty adjacent points) for Black Stone |
| | 4 | Number of liberties (empty adjacent points) for White Stone |
| Player color | 1 | Whether current player is black |
| Sensibleness | 1 | Whether a move at this point is a successful ladder escape |
| ko | 1 | Whether the point indicates ko |

Because the concept of liberty has important tactical significance for the game of Go, it needs to be modeled and encoded explicitly, and the same is true for ko. When the neural network model can directly see the properties of liberty, ko and the attribute of stone color, it can place more emphasis on these concepts in training, thus making it easier for the model to understand their impact on the game.

Based on the game records generated by this supervised learning neural network model, we train the model with 12 million weight parameters. The obtained model reaches an accuracy rate of 98% on the test set.

IV. DEMONSTRATION AND EXPERIMENTS

*A. System demonstration*

The CH-Go system is mainly developed in python3.7, and the demo page is developed in HTML and JavaScript. Fig. 5 represents the flow of data through the CH-Go system and the interface of the system is shown in fig. 6.

Players can use it from any device with a web browser, whether PC or mobile phone. The system uses the agent that iterates 20 times through reinforcement learning after completing the supervised learning. The system uses Chinese rules with no handicap and 7.5 komi, and players always play black. The system interface is simple and easy to use and we respect the privacy of players and will not collect any information from players, including moves and results of games.

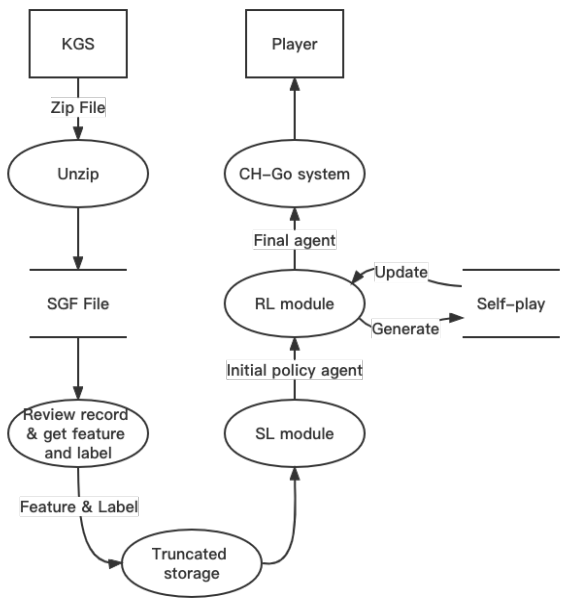

Fig. 5. Data Flow Diagram

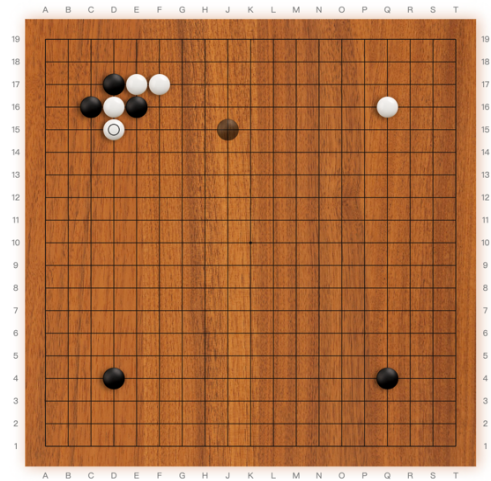

Fig. 7. The demonstration of escape

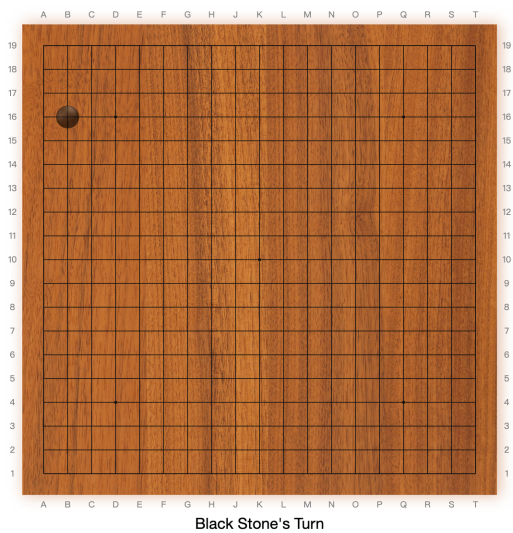

Fig. 6. System demonstration page

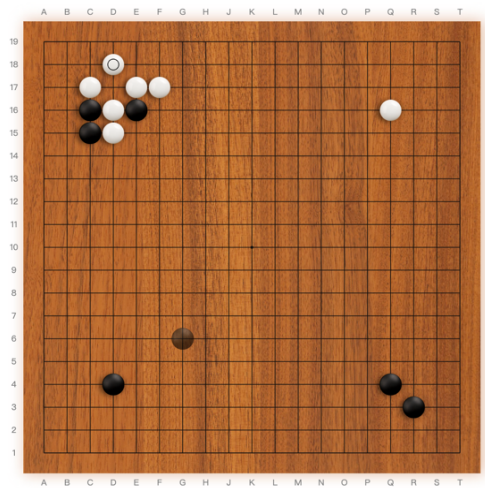

Fig. 8. The demonstration of capture

After visiting the page through the link, CH-Go can be tested. When a stone has only one liberty left, it will escape avoiding being captured. For example, as shown in Fig. 7, when Black moves to C16, White D16 has only one liberty left. At this time, CH-Go will choose to move D15 immediately to escape.

Later, when Black chooses to play C15, White will play C17 to avoid Black forming a ladder connection. CH-Go will also actively capture when the opponent makes a mistake to create an advantageous situation for itself, such as the Go game following Fig. 7; Black D17 has only one liberty left. If Black makes a mistake, not choosing to play D18 to escape, White will choose to capture, as shown in Fig. 8.

In addition, the clipping probability distribution ensures the diversity of the CH-Go's action choices. Even if Black always chooses the same move, CH-Go will still move in different places. As shown in Fig. 9, for the exact same Go move, CH-Go has different choices of moves.

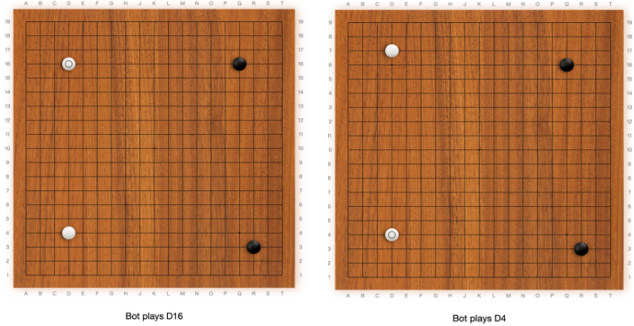

Fig. 9. Demonstration of system action selection diversity

In general, after completing supervised learning and reinforcement learning, CH-Go basically plays at a 3D amateur. In the early stage of the game, CH-Go is especially logical, and it plays fast, giving players a certain sense of pressure. In the middle game stage, the system will seize loopholes of opponents to fight to expand its advantages, and it will also have a sense of the overall situation when there is a local stalemate so that it deploys stones in other places, as shown in Fig. 10 for CH-Go and amateur 6D Go player's game indication:

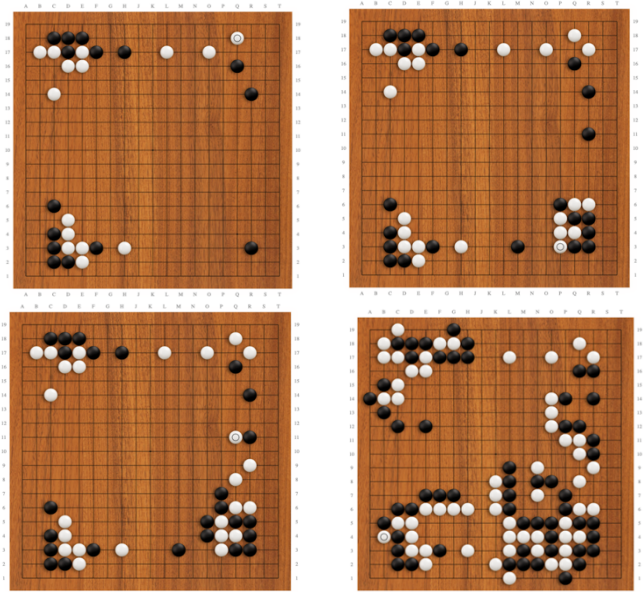

Fig. 10. CH-Go plays Go with a 6D Go player

## B. Experiments

To demonstrate the methodology validity and usability of the system, we compare CH-Go with the following baseline. a) Leela-zero is a Go program with no human-provided knowledge using Monte Carlo tree search (MCTS) and a deep residual convolutional neural network stack. b) I-Go is a growing man-machine system that determines the best step and completes playing decision by calculating Learning-Time Evaluation (LTE) and Playing-Time Evaluation (PTE) synthetically [1]. c) DPL-Go [18] is an implementation of algorithms based on deep reinforcement learning and Upper Confidence Bound Applied to Trees (UCT). Although there are some famous Go robots, AlphaGo, Fine Art, GOLAXY et cl., these are non-open, we cannot compare with them.

In terms of speed and computational complexity, Table 2 illustrates both speed and memory footprint results. Leela-Zero, a distributed open-source project running off of computation donated by volunteers has taken more than a year to reach top levels. It has 18 inputs to the first layer. Because of reducing feature planes as the input of the neural network, Zobrist-guided hash coding and truncated storage module, CH-Go has faster speed and smaller memory footprint.

TABLE II. COMPARISON BETWEEN CH-GO AND OTHER METHODS ABOUT METHODOLOGY VALIDITY

| Method | Training Time | The size of memory footprint |
|---|---|---|
| Leela-zero | A lot of GPU training more than a year | Generally larger |
| I-Go | None | None |
| DPL-Go | 60 hours | None |
| **CH-Go** | **50 hours** | **101.1MB** |

As illustrated in Table 3, in the aspect of usability, Players do not need to download and install CH-Go, and do not need to configure parameters.

TABLE III. COMPARISON BETWEEN CH-GO AND OTHER METHODS ABOUT USABILITY

| Method | Access via browser | Install | Configure parameters |
|---|---|---|---|
| Leela-zero | No | Yes | Yes |
| I-Go | No | Yes | No |
| DPL-Go | No | None | No |
| **CH-Go** | **Yes** | **No** | **No** |

In conclusion, the technique is useful for novices and regular amateurs. They have access to the browser whenever and whenever they want to play games with computers. Users can have a positive experience thanks to the page's simplicity, ease of use, and quick system response.

## V. CONCLUSION

Based on the 160k Go game records of KGS, we train the Go agent by mixing supervised learning and reinforcement learning, deploy it on the browser through the web front-end, lightweight Flask application and Amazon server, and establish a system for Go fans to access and play human-machine games anytime, anywhere. The main feature of this paper is that a truncated storage module is introduced in data processing, which stores a large amount of game data in blocks, accesses it on demand, and loads it in parallel, so as to solve the memory and speed problems caused by a large amount of Go data. At the same time, the speed of Go board games is accelerated by Zobrist-guided hash coding, the accuracy of supervised learning neural network is improved with our 11-feature-planes input representation, and the diversity and network stability of reinforcement learning action selection are improved by clipping probability distribution and random sampling.

There is still a lot of room for improvement. For example, the standard of the current system is not good enough. When encountering a strong opponent, it is easy to give up. Therefore, in the future we consider further improving the system by iterating more rounds of reinforcement learning and introducing the Monte Carlo tree search. In addition, we would like to enrich the problem setting, allowing the system to make the best choice even when the probability of winning is low. Go is a precise and vague, common and magical [19] game, and we expect a more robust and convenient Go system to be built in the future.

## REFERENCES

[1] Li Xin. The Study of Growing Man-Machine I-Go System [D]. Shanghai Jiao Tong University, 2009.

[2] Wang Yajie, Qiu Hongkun, Wu Yanyan, et al. Research and development of computer games[J]. CAAI Transactions on Intelligent Systems, 2016, 11(6): 788-798.

[3] Baudiš P, Gailly J. Pachi: State of the art open source Go program[C]//Advances in computer games. Springer, Berlin, Heidelberg, 2011: 24-38.

[4] Tian Y, Ma J, Gong Q, et al. Elf opengo: An analysis and open reimplementation of alphazero[C]//International Conference on Machine Learning. PMLR, 2019: 6244-6253.

[5] Tian Y, Zhu Y. Better computer go player with neural network and long-term prediction[J]. arXiv preprintarXiv:1511.06410, 2015.

[6] Silver D, Schrittwieser J, Simonyan K, et al. Mastering the game of go without human knowledge[J]. nature, 2017, 550(7676): 354-359.

[7] Silver D, Hubert T, Schrittwieser J, et al. Mastering chess and shogi by self-play with a general reinforcement learning algorithm[J]. arXiv preprint arXiv:1712.01815, 2017.

[8] Silver D, Huang A, Maddison C J, et al. Mastering the game of Go with deep neural networks and tree search[J]. nature, 2016, 529(7587): 484-489.

[9] Zobrist A L. A new hashing method with application for game playing[J]. ICGA Journal, 1990, 13(2): 69-73.

[10] Zeiler M D. Adadelta: an adaptive learning rate method[J]. arXiv preprint arXiv:1212.5701, 2012.

[11] Pumperla M, Ferguson K. Deep learning and the game of Go[M]. Manning Publications Company, 2019..

[12] Littman M L. Markov games as a framework for multi-agent reinforcement learning[M]//Machine learning proceedings 1994. Morgan Kaufmann, 1994: 157-163.

[13] Williams R J. Simple statistical gradient-following algorithms for connectionist reinforcement learning[J]. Machine learning, 1992, 8(3): 229-256.

[14] Clark C, Storkey A. Training deep convolutional neuralnetworks to play go[C]//International conference on ma- chine learning. PMLR, 2015: 1766-1774.

[15] Zhou Xing. Research and application of massive large, medium and small file storage system based on MongoDB [D]. China University of Geosciences (Beijing), 2016.

[16] Huang J, Zhang D, Miao H. The Research and Imple- mentation of Connect6 Intelligent Chess Game System[J]. Computer Knowledge and Technology, 2009, 5(25): 7198-7200.

[17] Gao Q, Guo C. Technology of hashing and its application research in hybrid game tree search engine of chinese chess[J]. Science Technology and Engineering, 2008, 8(17): 4869-4872.

[18] Deng Hangyu. The design and implementation of algorithms for board games based on deep reinforcement learning[D]. Nanjing University, 2018.

[19] Yue Peng. The study of algorithms in computer Go[D]. Southwest University, 2007